\documentclass{article}

\PassOptionsToPackage{numbers,compress}{natbib}

\usepackage[final]{neurips_fitml_2024}

\usepackage[utf8]{inputenc} 
\usepackage[T1]{fontenc}    
\usepackage[english]{babel}
\usepackage{csquotes}
\usepackage{hyperref}       
\usepackage{graphicx}
\usepackage{url}            
\usepackage{booktabs}       
\usepackage{amsfonts}       
\usepackage{nicefrac}       
\usepackage{microtype}      
\usepackage{xcolor}         
\usepackage{caption}
\usepackage{amsmath}
\usepackage{floatrow}
\newfloatcommand{capbtabbox}{table}[][\FBwidth]
\usepackage{blindtext}
\usepackage{listings} 
\usepackage{enumitem}

\title{Optimizing Small Language Models for In-Vehicle Function-Calling}

\author{%
Yahya Sowti Khiabani$^{1}$\thanks{Corresponding author: yahya.sowti@mercedes-benz.com} \quad Farris Atif$^{1}$ \quad Chieh Hsu$^{1}$ \quad \textbf{Sven Stahlmann}$^{2}$ \\ 
\quad \textbf{Tobias Michels}$^{2}$ \quad \textbf{Sebastian Kramer}$^{2}$ \quad \textbf{Benedikt Heidrich}$^{2}$ \\ 
\quad \textbf{M. Saquib Sarfraz}$^{2}$ \quad \textbf{Julian Merten}$^{1}$ \quad \textbf{Faezeh Tafazzoli}$^{1}$\\
$^{1}$Mercedes-Benz Research \& Development North America \\
$^{2}$Mercedes-Benz Tech Innovation
}

\begin{document}

\maketitle

\begin{abstract}
We propose a holistic approach for deploying Small Language Models (SLMs) as function-calling agents within vehicles as edge devices, offering a more flexible and robust alternative to traditional rule-based systems. By leveraging SLMs, we simplify vehicle control mechanisms and enhance the user experience. Given the in-vehicle hardware constraints, we apply state-of-the-art model compression techniques, including structured pruning, healing, and quantization, ensuring that the model fits within the resource limitations while maintaining acceptable performance. Our work focuses on optimizing a representative SLM, Microsoft's Phi-3 mini, and outlines best practices for enabling embedded models, including compression, task-specific fine-tuning, and vehicle integration. We demonstrate that, despite significant reduction in model size which removes up to 2 billion parameters from the original model, our approach preserves the model's ability to handle complex in-vehicle tasks accurately and efficiently. Furthermore, by executing the model in a lightweight runtime environment, we achieve a generation speed of 11 tokens per second, making real-time, on-device inference feasible without hardware acceleration. Our results demonstrate the potential of SLMs to transform vehicle control systems, enabling more intuitive interactions between users and their vehicles for an enhanced driving experience.

\end{abstract}

\section{Introduction}
The rapid evolution of vehicle software has created a complex ecosystem of interconnected Electronic Control Units (ECUs) via a Controller Area Network (CAN) bus. As consumer demand for advanced features like virtual voice assistants, interior functions, and ambient settings grows, seamlessly integrating these features into existing vehicle systems becomes increasingly complex. Traditionally, each new software feature requires extensive development to interface with core vehicle systems. Here, leveraging a SLM as intermediary to streamline communication between disparate systems may facilitate easier integration of new features and adjustments based on driver conditions or external software updates.

SLMs like Gemma (2B), Microsoft Phi-3 mini (3.8B), Mistral (7B), and Llama-3 (8B) have shown strong performance on academic benchmarks, despite being significantly smaller than traditional LLMs \cite{huggingface2024openllm}. However, due to the constraints of in-vehicle hardware, deploying these models directly may still be impractical. In this paper, we focused on optimizing these SLMs by further reducing their size and fine-tuning them to maintain performance on domain-specific tasks. We employed advanced model compression techniques such as pruning, quantization, and lightweight runtime execution.
Recent studies indicate that each of these techniques results in an acceptable performance loss when used individually. However, there is limited research on the combined use of these techniques while fine-tuning a model for specific tasks, such as function-calling.

Recently, \citet{patil2023gorillalargelanguagemodel} demonstrated the enhancement of function-calling capabilities in Llama-based models \cite{touvron2023llamaopenefficientfoundation} using a retrieval paradigm. Building on this, we extended their approach to the automotive domain. Specifically, we optimized a representative SLM to control various in-vehicle functions exposed via gRPC \cite{grpc}, such as seat heating, ambient lighting and more. This enables dynamic control of vehicle settings, reducing manual intervention and allowing seamless software updates.  

\begin{figure}[t]
    \centering
    \includegraphics[width=0.99\linewidth]{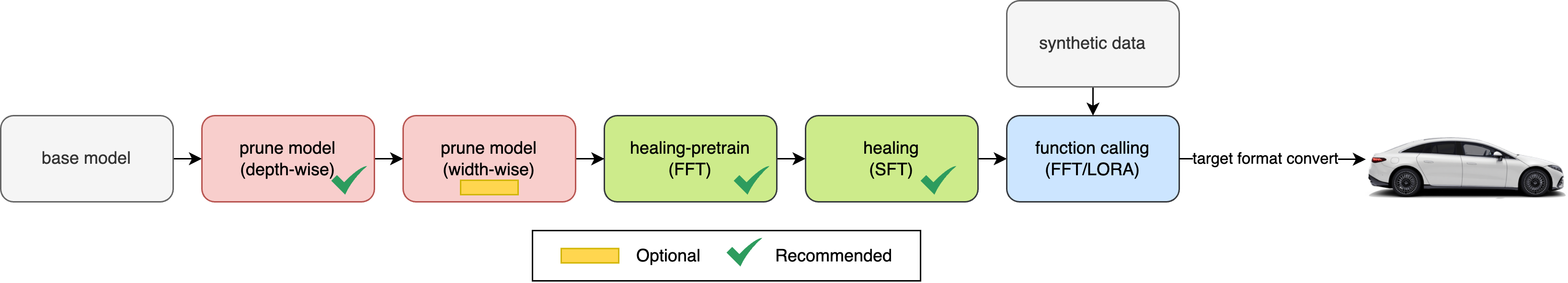}
    \caption{ \small Proposed framework for optimizing and deploying SLM for in-vehicle function-calling. Red represents the pruning stages, green for healing, and blue for function-calling alignment.}
    \label{fig:E2E-arch}
\end{figure}

Similar to function-calling capabilities, \citet{gromov2024unreasonable} showed that small models like Phi-2 could maintain baseline accuracy on reasoning tasks. Their approach used pruned decoder heads with a healing process using parameter efficient fine-tuning techniques like Low-Rank Adaptation (LoRA) \cite{hu2021lora}. For in-vehicle deployment, we found that a more robust healing process, including full fine-tuning, is necessary to maintain acceptable performance across benchmarks and real-world tasks. Furthermore, similar to the Octopus v2 approach \cite{chen2024octopusv2ondevicelanguage}, we introduced special tokens to represent function calls, aligning our pruned and healed model with in-vehicle function-calling tasks. The introduced special tokens map the model's token space to gRPC services, enabling the model to dynamically trigger specific vehicle settings. Figure \ref{fig:E2E-arch} presents an overview of our proposed framework.

In summary, our contributions are as follows:

\begin{enumerate}
    \item \textbf{Model Pruning and Healing}: We applied similarity-based depth-wise pruning, an optional width-wise pruning and healing techniques to reduce the size of the Phi-3 mini model while maintaining acceptable performance across both general and domain-specific tasks.
    \item \textbf{Fine-Tuning for In-Vehicle Function-Calling}: We fine-tuned the pruned and healed model using a custom dataset for in-vehicle function-calling, incorporating specialized tokens to map language model outputs to gRPC-based vehicle functions.
    \item \textbf{Efficient Deployment}: We leveraged llama.cpp for model conversion and quantization, enabling efficient deployment on resource-constrained automotive hardware. This ensures that the language model can operate in real-time environments with limited computational resources.
\end{enumerate}

The rest of the paper is organized as follows: Section~\ref{sec:related_work} presents the related works on pruning, healing, and function-calling. Section~\ref{sec:methods} details our methodology, including model pruning, healing, and the fine-tuning process for in-vehicle function-calling as well as model conversion and deployment strategies in vehicle. Finally, before concluding in Section~\ref{sec:conclusion}, Section~\ref{sec:results} presents our evaluation results.

\section{Related Work}
\label{sec:related_work}

\textbf{Model Pruning and Healing:} Model pruning reduces the size and complexity of machine learning models by removing less important weights, neurons, or entire layers \cite{lecun1989optimal}. The goal is to create a model with a reduced memory footprint, lower computational requirements, and faster inference. This is particularly useful for deploying models on resource-constrained devices, such as mobile phones or embedded systems like vehicle head units.

Pruning approaches can be broadly categorized into \textit{structured pruning} and \textit{unstructured pruning}. Structured pruning removes entire structures within the network, such as neurons, filters, or layers. In contrast, unstructured pruning removes individual weights regardless of their position in the network. While unstructured pruning can reduce the number of parameters, it often results in sparse networks, which are difficult to accelerate without custom hardware \cite{DBLP:journals/corr/LiKDSG16}, making it impractical for use in environments like vehicle head units.

In structured pruning, there are two primary approaches: \textit{depth-wise} and \textit{width-wise} pruning. Depth-wise pruning focuses on removing entire layers from the network. For example, ShortGPT \cite{men2024shortgpt} calculates a Block Influence metric on a calibration dataset and removes layers with the smallest scores. Another example is the layer collapse method \cite{yang2024laco}, which merges adjacent layers while ensuring that the output on a calibration dataset remains as close as possible to the original. \citet{gromov2024unreasonable} compute a similarity score between activations before and after a block of several layers and prune the block with the highest similarity.

Once pruning is performed, the model can experience degraded performance due to the abrupt removal of layers. To mitigate this, healing is applied through a small amount of fine-tuning to restore the model’s capabilities. As shown in \cite{gromov2024unreasonable}, after pruning layers based on activation similarity, LoRA is applied to fine-tune the remaining model on a small subset of the original training data. This step ensures that the pruned model recovers its performance across tasks. For example, Gromov et al. demonstrated that up to 45\% of layers, depending on the type of the model, could be pruned with minimal degradation in question-answering tasks when followed by this healing process.

Width-wise pruning aims to reduce the dimensionality of layers. For instance, SliceGPT \cite{ashkboos2023slicegpt} leverages the computational invariance of weight matrices to apply Principal Component Analysis (PCA) and identify rows and columns of the weight matrices that can be deleted. The MINITRON approach proposed by \citet{muralidharan2024compact} first records activation magnitudes for all hidden neurons and attention heads on a calibration dataset. Then, the neurons and attention heads with the lowest activation magnitude are pruned across all layers of the model.

Note that it is also possible to combine depth and width pruning strategies. For example, \citet{muralidharan2024compact} experimented with using their width pruning approach together with a depth pruning strategy similar to the one used by \citet{gromov2024unreasonable}.

\textbf{Function-Calling:} Function-calling is an emergent ability of language models which expands capabilities beyond text generation allowing them to interact with tools, APIs, and the physical world. Toolformer \cite{schick2023toolformer} showcased the ability of language models to use external tools through simple API calls, without explicit fine-tuning. Similarly, LangChain \cite{langchain} provides an interface for chain of thought with various tools and data sources. Moreover, retrieval-augmented generation (RAG) techniques, such as REALM \cite{guu2020realm}, have been shown to further enhance the accuracy and reliability of function-calling by leveraging external knowledge sources during inference.

As a pivotal progress in function-calling by small language models on edge devices, the Octopus v2 paper \cite{chen2024octopusv2ondevicelanguage} introduced a novel methodology by incorporating functional tokens directly into the tokenizer, thereby streamlining the function-calling process. This approach inspired our work, where we employed specialized MB tokens, akin to the functional tokens in Octopus v2, to map language model outputs to specific vehicle functions. We also adopted a strategy similar to Octopus v2's negative sample technique, incorporating irrelevant queries into our synthetic dataset to enhance the robustness of our model against unintended function activations.
\section{Methodology}
\label{sec:methods}
We selected Phi3-mini\footnote{\url{https://huggingface.co/microsoft/Phi-3-mini-4k-instruct/tree/ff07dc01615f8113924aed013115ab2abd32115b}. Note that we used the pre-July-update version of Phi3-mini.}, which is a decoder-only transformer language model with $L=32$ hidden layers, as a representative of an SLM due to its small size of 3.8B parameters, its strong performance across public benchmarks, and its ability to run across various software stacks \cite{abdin2024phi3technicalreporthighly}.

\subsection{Model Pruning}
We used both width and depth pruning in our experiments to produce two different variants of the original Phi3-mini: Phi3-2.8B and Phi3-1.8B. Table~\ref{tab:model_details} contains details regarding the architecture of the two variants and the original Phi3-mini.

\begin{figure}[t]
\centering
\includegraphics[width=0.7\textwidth]{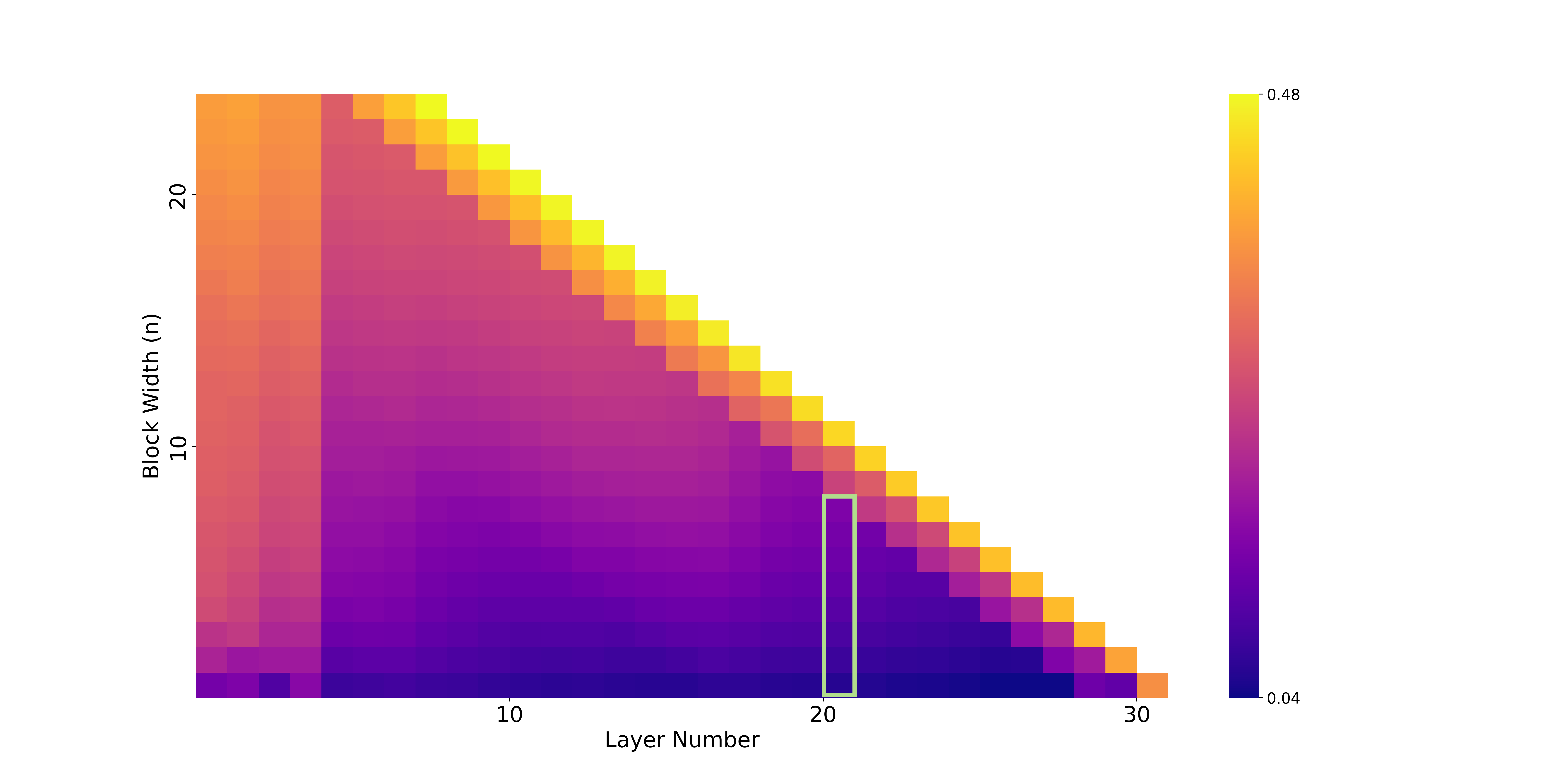}
\caption{Heatmap of distances for all 32 decoder layers of Phi-3, with varying block sizes \(n \in \{1,\ldots, 24\}\), calibrated with the fineweb dataset. Dark purple indicates regions of minimum distance or maximum similarity. Layers 21-29 (highlighted in green) were found to be the optimal block of size \(n=8\) to prune.}
\label{fig:block_sim}
\end{figure}

For depthwise pruning, we pruned a contiguous block of size $n=8$ which minimized cumulative layer distance between decoder layers. Here we followed general guidance from \citet{gromov2024unreasonable}, where it was noted that dropping more than 30\% of the layers across different model families (Llama, Qwen, Mistral, Phi2, etc) resulted in collapse of the underlying model. Let $h_i$ represent the $i$-th hidden state of the model and $n$ the chosen block size. Then, for all $i \in \{1, \ldots, L - n\}$, where $L$ is the number of hidden layers in the model, we computed the angular distance between hidden states as $d(h_{i}, h_{i+n}) := \arccos\left(\frac{\langle h_i, h_{i+n} \rangle}{\lVert h_i \rVert \lVert h_{i+n} \rVert}\right)$. Figure~\ref{fig:block_sim} shows the resulting distances for different block sizes calculated against the fineweb dataset \citep{penedo2024finewebdatasetsdecantingweb}. Layers 21-29 were maximally similar, and thus pruned. The resulting model is denoted as Phi3-2.8B.

 Phi3-1.8B was then created by applying the width pruning approach from \citet{muralidharan2024compact} to Phi3-2.8B by recording activations on each layer (block) in the same manner as the depth-wise approach. Next, for the attention heads, we calculated the L2 norm along the head dimension and computed the mean across the sequence and batch dimensions for all activations. This gives us a score for each hidden neuron, each neuron in the intermediate layer of the multi-layer perceptron (MLP), and each attention head. Finally, we pruned the neurons and attention heads with the lowest score: the hidden dimension from 3072 to 2688, the MLP dimension from 8192 to 5120, and the number of attention heads from 32 to 28.

\begin{table}[t]
\centering
\small
    \begin{tabular}{lccccc}
    \hline
    \textbf{Model} & \textbf{Parameters} & \textbf{\# Layers} & \textbf{Hidden dim} & \textbf{MLP dim} & \textbf{Attention heads} \\ \hline
    Phi3-mini & 3.8B  & 32 & 3072 & 8192 & 32 \\
    Phi3-2.8B & 2.8B  & 24 & 3072 & 8192 & 32  \\
    Phi3-1.8B & 1.8B  & 24 & 2688 & 5120 & 28\\ \hline
    \end{tabular}%
\caption{Model architecture details for original phi3-mini and its pruned variants.}
\label{tab:model_details}
\end{table}

\subsection{Healing and Instruction Tuning}\label{sec:healing}

After pruning, the resulting models struggle to generate coherent sentences and lose their alignment.
As described in \citet{gromov2024unreasonable}, this can be remedied by applying a healing training afterwards.
The authors suggest training for 5000 steps using QLoRA \citep{dettmers2023qlora} on only the MLP weights with a diverse web-scale dataset, for which we used fineweb dataset.
We denote the models produced by this step with $h_\text{short}$.

However, we determined that this was not sufficient to fully recuperate the model.
While the resulting models were able to form correct and meaningful sentences again, the factual knowledge of the original Phi3-mini was almost entirely lost.
Moreover, we observed that the healing loss was still decreasing. As a result, we continued training the pruned models on fineweb-edu \citep{lozhkov2024fineweb-edu} for another 45000 steps/~15B tokens, denoting the result as $h_\text{long}$.
This roughly matches the approach from \citet{muralidharan2024compact} who heal their pruned models for 10B tokens.

Afterwards, we instruction tuned the models for one epoch on the OpenHermes-2.5 dataset \citep{OpenHermes2.5}. The resulting models are marked with \emph{SFT} (Supervised Fine-Tuning). We will further discuss the importance of longer fine-tuning and instruction-tuning (Phi3-2.8B + $h_\text{long}$ + \emph{SFT}) on getting noticeable improvement on the quality of responses generated by the model in results presented in section~\ref{sec:results}.

For the Phi3-1.8B model, note that we first applied healing to the pruned Phi3-2.8B model before the width pruning step, i.e., we used Phi3-2.8B + $h_\text{long}$ as the base model and then applied width pruning and instruction fine-tuning (SFT) on top of that.

\subsection{Fine-tuning for In-Vehicle Function-Calling}

Fine-tuning language models has become a standard practice, with various approaches being explored. Both full fine-tuning (FFT) and LoRA \citep{hu2021loralowrankadaptationlarge} are widely used methods, each with its own strengths and weaknesses. FFT offers comprehensive model adaptation but can be computationally expensive, while LoRA provides a more parameter-efficient alternative, particularly beneficial when GPU resources are limited. In our research, we leveraged both full model training and LoRA training, allowing us to compare their performance and understand their trade-offs. LoRA's ability to extend model functionalities further highlights its potential for adapting our framework to a broader range of applications. In addition to being more computationally efficient, the modularity of LoRA adapters opens up the possibility of seamlessly switching between different adapters, allowing for dynamic customization and adaptation of the model to various tasks or domains, as explored in works like LoRA-Switch \cite{kong2024loraswitchboostingefficiencydynamic}.
Building on the methodology of Octopus v2 \cite{chen2024octopusv2ondevicelanguage}, we fine-tuned the pruned and healed phi3-mini model to enhance its function-calling capabilities for in-vehicle operations.

\textbf{Synthetic Dataset Generation for Fine-Tuning:} We generated a comprehensive synthetic dataset inspired by the Octopus v2 model's technique of integrating functional tokens into the tokenizer. Eight \texttt{MB} tokens were defined for specific vehicle functions, such as \texttt{set\_ambient\_light\_color\_program} mapped to \texttt{<MB\_1>} and \texttt{set\_seat\_heating\_intensity} mapped to \texttt{<MB\_2>}. To ensure both diversity and naturalness, we utilized a multi-step prompt design for generating positive and negative examples.

\textbf{Positive Examples:} We used a prompt template to generate 25,000 examples evenly distributed across all vehicle functions. For example:
\begin{verbatim}
Query:
Warm up my seat and set the mood to Malibu Sunset before I get in the car
Response:
<MB_2>(seat_position="FRONT_LEFT", intensity=3);
<MB_1>(color_program="MalibuSunset");
<MB_0>(message="I've warmed up your seat and set the ambient lighting 
to Malibu Sunset. Your car will be inviting when you get in.")<MB_end>
\end{verbatim}

\textbf{Negative Examples:} To improve model robustness, 500 irrelevant queries were generated using a negative sampling strategy. These queries were plausible but unsolvable by the provided functions (e.g., "Can you teleport the car to Hawaii?"). The assistant responds by politely declining the request.

\textbf{Quality Control:} To ensure the generated dataset reflects real-life spoken user queries, we manually curated a subset of examples derived from common user questions and included them in the prompt to the LLM. We enforced an even distribution of function calls across different functions to avoid imbalance. Specific rules were added to the prompts to ensure high-quality dataset generation, and this, combined with de-duplication and post-processing, maintained a high standard for the final examples.

The dataset thus reflects natural in-vehicle commands, ensuring both accuracy in function-calling and robustness to unsupported queries. Examples of the functions vehicle services are provided in the supplementary materials.

\textbf{Fine-Tuning Settings:} We fine-tuned both the 2.8B and 1.8B pruned models using LoRA fine-tuning and full fine-tuning, with the specific settings outlined in Table \ref{tab:finetuning-settings}. Additionally, we fine-tuned the original Phi-3 Mini model using LoRA for comparison purposes. It is important to note that for FFT, we limited the training to just one epoch and used a smaller learning rate along with a weight decay of 0.1 as a form of regularization \cite{DBLP:journals/corr/abs-1711-05101}. This approach aimed to prevent overfitting to the function-calling dataset, which is a common concern with FFT due to its tendency to aggressively adapt to the training data. 

In contrast, LoRA fine-tuning required at least two epochs of training without any weight decay and with a larger learning rate to achieve satisfactory results on function-calling tests. This difference in training regimes can be attributed to the nature of LoRA, which introduces a smaller set of trainable parameters compared to FFT, and that is why it may necessitate more training epochs and a higher learning rate to effectively capture the nuances of the function-calling task.

\begin{table}[h]
\centering
\caption{Fine-tuning settings for LoRA and FFT}
\label{tab:finetuning-settings}
\footnotesize 
\begin{tabular}{lcc}
\hline
\textbf{Parameter} & \textbf{LoRA} & \textbf{FFT} \\ \hline
\textbf{Rank}             & 96   & -   \\ 
\textbf{Alpha}            & 16   & -   \\ 
\textbf{Batch Size}       & 4    & 4   \\ 
\textbf{Learning Rate}    & 5e-5 & 5e-6 \\ 
\textbf{Number of Epochs} & 2    & 1    \\ 
\textbf{Weight Decay}     & 0  & 0.1  \\ 
\textbf{Target Modules} & 
\begin{tabular}[c]{@{}l@{}}embed\_tokens, qkv\_proj,  o\_proj, mlp (up/down), lm\_head\end{tabular} &
- \\ \hline
\end{tabular}
\end{table}

\subsection{Model Deployment}

There are various lightweight libraries optimized for on-device inference, such as:
MLX, tensorflow-lite, ONNX, and ExecuTorch \cite{mlx2023}\cite{bai2019}\cite{tensorflow2015-whitepaper}\cite{executorch}. We chose to leverage the open source on-device runtime, llama.cpp, to host our resulting pruned Phi-3 model.

llama.cpp is a wrapper around the ggml tensor library, which has native support for transformer model operations \cite{ggerganov2023llama}. The framework makes use of the gguf file format to serialize language models and respective metadata (tokenizer, model type, quantization, etc.) into a single artifact, which is then executed against the ggml tensor library. It is flexible in its implementation and operations can be removed or composed depending on the model graph being executed.

Specifically, we took the following steps to convert our model into the target format: merge LoRA into HF base model (If LoRA is used), convert safetensors artifact to gguf, quantize resulting gguf to 4-bit, test resulting artifact, and quantify distance between gguf and original safetensors implementation.

While gguf artifacts can be quantized from 2-bit to 8-bit, we chose a 4-bit quantization strategy. In doing so, we balanced token throughput and generation with minimal added perplexity. 
Additionally, in this format a pruned Phi3 model uses less than 2gb of RAM. 

\section{Experimental Results}
\label{sec:results}

\subsection{Evaluation Results on the Original Format} \label{ssec:mbti_benchmarks}

To assess the model's general language understanding, after various pruning, healing, and training steps, we used the \texttt{lm-evaluation-harness}\footnote{\url{https://github.com/EleutherAI/lm-evaluation-harness}} framework to evaluate models in their original safetensor format. The evaluation covered multiple benchmarks, including standard tasks such as question answering, natural language understanding, and reasoning. Specifically we evaluated our models on Winogrande \cite{sakaguchi2019winograndeadversarialwinogradschema}, TruthfulQA \cite{lin2022truthfulqameasuringmodelsmimic}, MMLU \cite{hendrycks2021measuringmassivemultitasklanguage}, HellaSwag \cite{zellers2019hellaswagmachinereallyfinish}, and ARC \cite{chollet2019measureintelligence}. The results are outlined in Table \ref{tab:bench_results}.
  
\begin{table}[t]
    \centering
    \caption{Benchmark results across model variations}
    \label{tab:bench_results}
    \resizebox{\textwidth}{!}{%
    \begin{tabular}{llllllll}
        \toprule
        Model & Winogrande & TruthfulQA & MMLU & HellaSwag & ARC & Avg \\
        \midrule
        Phi3-3.8B & 0.74 & 0.36 & 0.70 & 0.59 & 0.54 & 0.59 \\
        Phi3-2.8B + $h_\text{short}$ & 0.69 & 0.34 & 0.65 & 0.47 & 0.42 & 0.51 \\
        Phi3-2.8B + $h_\text{short}$ + \emph{SFT} & 0.71 & 0.29 & 0.55 & 0.49 & 0.41 & 0.49 \\
        Phi3-2.8B + $h_\text{long}$ + \emph{SFT} & 0.68 & 0.27 & 0.56 & 0.51 & 0.46 & 0.50 \\
        Phi3-1.8B + $h_\text{long}$ + \emph{SFT} & 0.62 & 0.27 & 0.42 & 0.44 & 0.36 & 0.42 \\
        Phi3-3.8B + $\emph{LoRA}$ (4-bit)  & 0.72 & 0.33 & 0.66 & 0.57 & 0.53 & 0.56 \\
        Phi3-2.8B + $h_\text{long}$ + \emph{SFT} + \emph{LoRA} (4-bit)  & 0.67 & 0.25 & 0.50 & 0.48 & 0.41 & 0.46 \\
        Phi3-2.8B + $h_\text{long}$ + \emph{SFT} + \emph{FFT} (4-bit)  & 0.66 & 0.26 & 0.51 & 0.47 & 0.41 & 0.46 \\
        Phi3-1.8B + $h_\text{long}$ + \emph{SFT} + \emph{LoRA} (4-bit)  & 0.60 & 0.28 & 0.32 & 0.41 & 0.34 & 0.39 \\
        Phi3-1.8B + $h_\text{long}$ + \emph{SFT} + \emph{FFT} (4-bit)  & 0.60 & 0.28 & 0.35 & 0.41 & 0.34 & 0.40 \\
    \bottomrule
    \end{tabular}%
    }
\end{table}

\textbf{Impact of Pruning and Healing:} Depth-wise pruning led to a modest decline in model performance - considering that approximately 1B parameters were removed. Moreover it is seen that longer healing yields better scores on MMLU, HellaSwag, and ARC for (Phi3-2.8B + $h_\text{long}$) vs (Phi3-2.8B + $h_\text{short}$). However, width-wise pruning on top of the 2.8B model caused significant degradation in model capabilities across all benchmarks regardless of the healing and alignment strategy applied to it. As a result, it can be inferred that regardless of the healing strategy applied, the upper limit of parameter removal is roughly 30\% of the original model parameters. This falls in-line with the findings from \citet{gromov2024unreasonable}.

\textbf{Impact of Instruction Tuning:} Instruction tuning following healing improves performance in some benchmarks (e.g., Winogrande up to 0.71 for $h_{short} + SFT$), though the overall average score remains close to 0.50 for most configurations. This suggests that instruction tuning can partially recover performance in certain tasks, although the improvements are task-specific and not universal across all benchmarks.

Although the benchmark results in Table~\ref{tab:bench_results} indicate that longer training and instruction tuning ($h_\text{long}$ + \emph{SFT}) lead to only slight improvements in benchmark performance, direct interaction with the model revealed noticeable enhancements compared to shorter training with QLoRA ($h_\text{short}$ + \emph{SFT}). 
We observed that, compared to the depth-pruned 2.8B Phi3 Mini model, the combination of long healing together with instruction tuning denoted as $h_\text{long}$ + \emph{SFT}, helps recover some of the performance lost due to pruning across various datasets. This is why we selected this configuration as the baseline for the next step, which involves further fine-tuning for function-calling.

\textbf{Impact of Function-Calling Fine-Tuning:} We evaluated the fine-tuned models with function-calling dataset on these benchmarks after 4-bit quantization denoted as (4-bit). Notably, there was no significant performance drop when comparing the phi3-3.8B model to its LoRA fine-tuned counterpart, phi3-3.8B + \emph{LoRA}. Similarly, across different benchmarks, the transition from $h_\text{long}$ + \emph{SFT} to $h_\text{long}$ + \emph{SFT} + (\emph{FFT} / \emph{LoRA}) did not result in substantial degradation for both the phi3-2.8B and phi3-1.8B models.

\subsection{Evaluation on Target Format}

Since the models will run with llama.cpp in vehicle, it is imperative to check performance after conversion from safetensor to 4-bit gguf. Unfortunately, the \texttt{lm-evaluation-harness} cannot interface with models in this format.
So, we used llama.cpp's own evaluation tool for running standard benchmarks on the converted models. Additionally, to evaluate model performance on function-calling tasks, we used an exact match metric which measures accuracy for both the function name and arguments. Table~\ref{tab:bench_results_after_ft} presents the results of evaluations on 4-bit gguf models.

\begin{table}[t]
    \centering
    \small  
    \caption{Benchmark results across model variations (4bit gguf)}
    \label{tab:bench_results_after_ft}
    \begin{tabular}{lllll} 
    \toprule
    Model & function-calling & TruthfulQA         & MMLU               & HellaSwag          \\ \midrule
    Phi3-3.8B + LoRA                        & 0.86 & 32.44 & 39.10 & 74.82 \\
    Phi3-2.8B + $h_{\text{long}}$ + \textsc{SFT}      & - & 26.81 & 34.51 & 65.86 \\
    Phi3-2.8B + $h_{\text{long}}$ + \textsc{SFT} + \textsc{FFT} & 0.88 & 26.57 & 34.30 & 63.05 \\
    Phi3-2.8B + $h_{\text{long}}$ + \textsc{SFT} + \textsc{LoRA} & 0.88 & 25.70 & 34.15 & 63.93 \\
    Phi3-1.8B + $h_{\text{long}}$ + \textsc{SFT}        & - & 26.32 & 31.18 & 55.54 \\
    Phi3-1.8B + $h_{\text{long}}$ + \textsc{SFT} + \textsc{FFT}  & 0.84 & 25.70 & 30.78 & 54.47 \\
    Phi3-1.8B + $h_{\text{long}}$ + \textsc{SFT} + \textsc{LoRA} & 0.86 & 27.54 & 31.03 & 54.65 \\ \bottomrule
    \end{tabular}
\end{table}

\textbf{Results on General Language Understanding:} The benchmark results across TruthfulQA, MMLU, and HellaSwag reveal a clear correlation between model size and performance on general language understanding tasks. As the model size decreases from 3.8B to 2.8B to 1.8B parameters, there's a consistent decline in performance across all three benchmarks. The data indicates that reducing the model size from 3.8B to 2.8B results in a performance drop of about 4.5 to 5 points on MMLU, while further reducing to 1.8B leads to an additional drop of approximately 3 points.

However, within models of the same size (e.g., Phi3-2.8B and Phi3-1.8B variants), the performance differences before and after fine-tuning are minimal. For example, Phi3-2.8B + SFT achieves an MMLU score of 34.51, while its fine-tuned variant, Phi3-2.8B + SFT + FFT, scores 34.30, showing little impact from function-calling fine-tuning. This suggests that the fine-tuning process does not negatively affect general language understanding and is applied effectively without diminishing the models’ performance on these benchmarks.

\textbf{Results on Function-Calling Performance:} The fine-tuning process leads to considerable improvements in function-calling accuracy across all models. Fine-tuning the Phi3-2.8B model with LoRA achieves a function-calling accuracy of 0.88 similar to FFT score. Even in the case of the smallest model, Phi3-1.8B, function-calling accuracy remains high, with LoRA achieving 0.86 and FFT scoring 0.84. These results highlight the efficiency of fine-tuning in enhancing task-specific performance, particularly for function-calling in vehicular systems. The minor variations in function-calling accuracy among models of different sizes suggests that even smaller models, such as Phi3-1.8B, can effectively handle specialized tasks like function-calling. Considering the current production speech system for function calling, which achieves an accuracy of 0.75 across various vehicle functions, the function-calling accuracy of 0.85 or higher demonstrated by the SLM presents a promising improvement over the existing system.

It should be noted that the message in <MB\_0> is not evaluated in function-calling performance since the ability for general language understanding was already evaluated. Although it degraded after fine-tuning for in-vehicle function-calling, the message remains fluent and explains what functions are activated based on human evaluation. The following example is from Phi3-2.8B $h_{long} + SFT + Lora FT$ (4-bit) which clearly depicts how a function activation addresses user needs:

\begin{verbatim}
Query: 
It's stuffy here, can you do something?
Response:
<MB_4>(zone="FRONT_LEFT", fan_value=3);
<MB_0>(message="I'm increasing the fan speed to help circulate 
air in your area.")
\end{verbatim}

\subsection{Token Generation Performance Gains} 

\begin{figure}[t]
    \centering
    \includegraphics[width=0.80\linewidth]{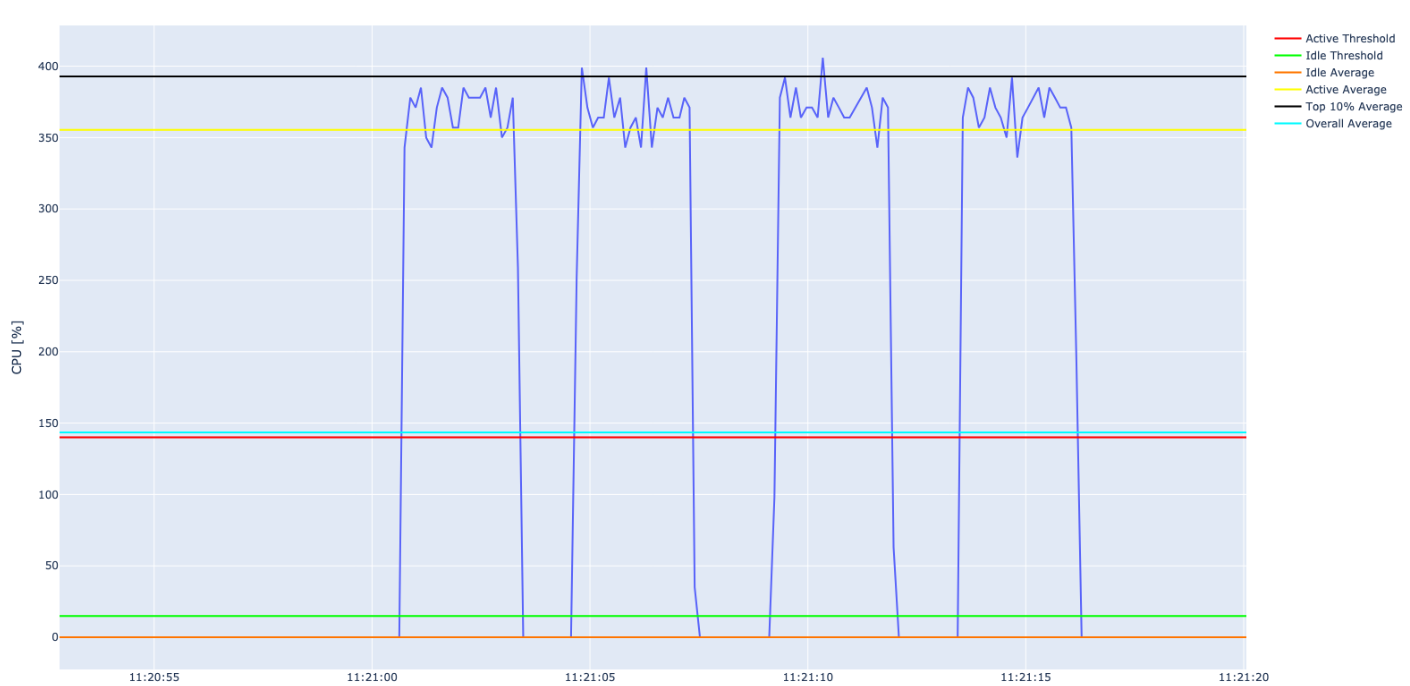}
    \caption{ \small CPU usage of LLM process during inference on the vehicle head unit. The horizontal lines show binned values of the process across time. The \textit{Top 10\% average}  (black line) shows the top 10\% of CPU usage of the process.}
    \label{fig:cpu}
\end{figure}

Table \ref{tab:bench_results_tps} summarizes the token generation results of Phi3-mini across all experiments. It is worth noting that depth-wise pruning yields a 2x increase in token-generation vs width-wise pruning. It can be inferred that removing decoder blocks altogether is more consequential in the model's ability to generate high quality responses, as well as generation speed. Additionally, the 1.8B parameter model achieves a token generation speed of 11 tokens/sec on CPU. For reference, a Llama model running on an NPU achieves the same performance \cite{qualcomm2023}. Figure \ref{fig:cpu} shows the CPU usage of the 1.8B model on a vehicle head unit. At inference, the pruned model uses 400\% CPU (the underlying CPU is an ARM processor with 7 cores). It can be inferred then that even a standard Phi3 which is typically regarded as a \textit{small} language model would use all available cores during inference, further demonstrating the need for a pruning step pre-deployment. While the spike is significant, it is only sustained for the duration of inference and can be further mitigated by dynamically allocating resources to the LLM process before inference. Moreover given the magnitude of the model and applications which it can unlock, the tradeoff is reasonable. As a result we demonstrate the benefit and feasibility of running smaller language models on-device, for an out-of-distribution use case (function-calling) without any hardware acceleration.

\begin{table}[t]
    \centering
    \caption{Benchmark results for 4-bit gguf model (tokens per second)}
    \label{tab:bench_results_tps}
    \small  
    \begin{tabular}{ll}
        \toprule
        \textbf{Model} & \textbf{t/s} \\
        \midrule
        Phi3 (base)     & 6.76  \\
        Phi3 (2.8B)     & 9.44  \\
        Phi3 (1.8B)     & 11.02 \\
        \bottomrule
    \end{tabular}
\end{table}

\section{Conclusion}
\label{sec:conclusion}

This work demonstrates the effective optimization of Small Language Models (SLMs) for in-vehicle function-calling, delivering high task accuracy and real-time performance on resource-constrained automotive hardware. Through structured pruning, healing, and fine-tuning, we significantly reduced the size of the Phi-3 mini model while preserving its ability to handle both general language tasks and specific vehicle functions. Our method shows that pruned and quantized models can efficiently perform real-time function execution, generating up to 11 tokens per second without hardware acceleration. This offers a scalable, flexible solution for modern vehicle control systems, enabling more intuitive user interactions. Future work can focus on enhancing general language understanding and further refining these models for specific automotive tasks.

\bibliography{references}


\clearpage
\begingroup
\fontfamily{lmss}\selectfont
\section*{Supplementary Material}
\label{sec:supplementary}

\lstset{
  language=Python,
  keywordstyle=\color{blue}, 
  basicstyle=\ttfamily\footnotesize, 
}

\textcolor{blue}{\textbf{Vehicle function examples:}}

\begin{lstlisting}
def set_ambient_light_color_program(color_program: str):
    """
    Set ambient light program in the car.

    Parameters:
    - colorProgram (str): Color programs options are 
    ["OceanBlue", "MiamiRose", "MalibuSunset", "BurningBlue", 
    "VenicePink", "ChromeShine", "RedMoon", "JungleGreen", 
    "Ultramarin", "FreshCyan", "RacingYellow", "RacingRed", 
    "AmethystHeat", "RoseGoldSparkle"]
    """
\end{lstlisting}

\begin{lstlisting}
def set_seat_heating_intensity(seat_position: str, intensity: int):
    """
    Set seat heating intensity in the car.

    Parameters:
    - seatPosition (str): Seat position options are 
    ["FRONT_LEFT", "FRONT_RIGHT", "REAR_LEFT", "REAR_RIGHT"]
    - intensity (str): Intensity options are [0, 1, 2, 3]
    """
\end{lstlisting}

\begin{lstlisting}
def set_temperature(zone: str, temperature: float, unit: str):
    """
    Set temperature in the car.

    Parameters:
    - zone (str): Zone options are 
    ["FRONT_LEFT", "FRONT_RIGHT", "REAR_LEFT", "REAR_RIGHT"]
    - temperature (float): Temperature range is 
    from 60 to 84 for FAHRENHEIT and 16 to 28 for CELSIUS
    - unit (str): Unit options are ["CELSIUS", "FAHRENHEIT"]
    """
\end{lstlisting}

\begin{lstlisting}
def set_window_position(window_position: str, operation: str):
    """
    Set window position in the car.

    Parameters:
    - windowPosition (str): Window position options are 
    ["FRONT_LEFT", "FRONT_RIGHT", "REAR_LEFT", "REAR_RIGHT"]
    - operation (str): Operation options are ["OPEN", "CLOSE"]
    """
\end{lstlisting}

\begin{lstlisting}
def respond_chat(message: str):
    """
    Respond to the user's query, for example to 
    provide an answer or ask for more information.

    Parameters:
    - message (str): The message that should be returned to the user.
    """
\end{lstlisting}

\end{document}